\documentclass{article}

\usepackage{PRIMEarxiv}

\usepackage[utf8]{inputenc} 
\usepackage[T1]{fontenc}    
\usepackage{hyperref}       
\usepackage{url}            
\usepackage{booktabs}       
\usepackage{amsfonts}       
\usepackage{nicefrac}       
\usepackage{microtype}      
\usepackage{lipsum}
\usepackage{amssymb}
\usepackage{fancyhdr}       
\usepackage{footmisc}
\usepackage{graphicx}       
\usepackage{authblk}
\graphicspath{{media/}}     

\title{Large Language Multimodal Models for 5-Year Chronic Disease Cohort Prediction Using EHR Data}

\author[1,*]{Jun-En Ding}
\author[2,*]{Phan Nguyen Minh Thao}
\author[2]{Wen-Chih Peng}
\author[2]{Jian-Zhe Wang}
\author[2]{Chun-Cheng Chug}
\author[2]{Min-Chen Hsieh}
\author[2]{Yun-Chien Tseng}
\author[3]{Ling Chen}
\author[4]{Dongsheng Luo}
\author[7]{Chenwei Wu}
\author[5]{Chi-Te Wang}
\author[5]{Pei-fu Chen}
\author[1]{Feng Liu}
\author[6,$\dagger$]{Fang-Ming Hung}

\affil[1]{School of Systems and Enterprises, Stevens Institute of Technology, Hoboken, USA}
\affil[2]{Department of Computer Science, National Yang Ming Chiao Tung University, Hsinchu City, Taiwan}
\affil[3]{Institute of Hospital and Health Care Administration, National Yang Ming Chiao Tung University, Taipei City, Taiwan}
\affil[4]{School of Computing and Information Science, Florida International University, USA}
\affil[5]{Center of Artificial Intelligence, Far Eastern Memorial Hospital, New Taipei City, Taiwan}
\affil[6]{Surgical Trauma Intensive Care Unit, Far Eastern Memorial Hospital, New Taipei City, Taiwan}
\affil[7]{Electrical Engineering and Computer Science, University of Michigan, MI, USA}
\affil[*]{These authors contributed equally to this work}
\affil[$\dagger$]{Corresponding author}

\begin{document}
\maketitle

\begin{abstract}
Chronic diseases such as diabetes are the leading causes of morbidity and mortality worldwide. Numerous research studies have been attempted with various deep learning models in diagnosis. However, most previous studies had certain limitations, including using publicly available datasets (e.g. MIMIC), and imbalanced data. In this study, we collected five-year electronic health records (EHRs) from the Taiwan hospital database, including 1,420,596 clinical notes, 387,392 laboratory test results, and more than 1,505 laboratory test items, focusing on research pre-training large language models. We proposed a novel Large Language Multimodal Models (LLMMs) framework incorporating multimodal data from clinical notes and laboratory test results for the prediction of chronic disease risk. Our method combined a text embedding encoder and multi-head attention layer to learn laboratory test values, utilizing a deep neural network (DNN) module to merge blood features with chronic disease semantics into a latent space. In our experiments, we observe that clinicalBERT and BiomedBERT, when combined with attention fusion, can achieve an accuracy of 73\% in multiclass chronic diseases and diabetes prediction. By transforming laboratory test values into textual descriptions and employing the Flan T-5 model, we achieved a 76\% Area Under the ROC Curve (AUROC), demonstrating the effectiveness of leveraging numerical text data for training and inference in language models. This approach significantly improves the accuracy of early-stage diabetes prediction.
\end{abstract}


\section{Introduction}

Chronic diseases are a major medical problem around the world. According to the WHO \cite{who-2020}, human mortality from chronic diseases is increasing. More than 70\% of the patient's income is spent on treating this disease. This is not only an individual health problem, but also an important public health issue that puts society at risk. Electronic Health Records (EHRs) collected the information needed to make these medical decisions across a variety of records, including the patient's medical history, laboratory test results, and imaging reports. As part of medical practice, all this information is incorporated into doctor's notes to document and summarize patient care.

In recent years, more and more researchers have utilized machine learning (ML) methods for the diagnosis of chronic diseases. Doctors face several challenges when using machine learning methods for diagnostic predictions, particularly when dealing with the intricacies of extracting data from EHRs data including numerical values, categories, and other formats. This will make it difficult for basic machine learning methods to handle medical terminology, complex sentence structures, ambiguity, uncertainty, contextual understanding. These challenges will necessitate significant human effort \cite{Bisercic-2023}. 

Although numerous natural language processing (NLP) models have been developed over the past few years to predict disease in clinical notes, the presence of tabular EHRs makes it difficult to rely solely on one model for text classification \cite{Mullenbach-2018}. In recent years, Large Language Models (LLMs) have begun to be trained successfully in a large number of corpus and have shown significant effectiveness in natural language processing tasks \cite{Zhao-2023}. These models undergo training in large volumes of textual data, enabling them to discern intricate statistical relationships embedded within words and phrases. Furthermore, researchers have begun to combine modality data with the LLM model \cite{Wu-2023}. This addresses the complexities of data extraction and the challenges associated with textual modeling utilizing tabular data. Examples of such applications include text classification \cite{Gasparetto-2022, Sun-2023} and even extend into the realm of clinical prediction within the intricate landscape of the medical field \cite{Zhang-2023, Steinberg-2021}. 

However, numerous medical studies in LLMs face constraints arising from the restricted availability of clinical notes corpus samples \cite{belyaeva2023multimodal}, such as the MIMIC or UK Biobank dataset \cite{bycroft2018uk}, or from inherent imbalances possibly related to specific diseases \cite{Minot-2022}. This will lead to biases in the predictive capabilities and usability of LLMs in clinical settings. 

 In our study, we propose a novel Large Language Multi-Modals (LLMMs) framework that can simultaneously integrate five years of textual information from EHRs and patients' laboratory test values for disease risk prediction. 

The key contributions of our work are as follows:
\begin{itemize}
    \item This is the first study to collect five years' worth of EHRs data and laboratory test values with the aim of utilizing LLMs for predicting chronic diseases, especially diabetes.
    \item We investigate the conversion of laboratory test values into textual information for training Large Language Models (LLMs). This method offers advantages in addressing missing patient laboratory test values and in facilitating more effective contextual learning for LLMs.
    \item Our proposed method demonstrates superior performance compared to state-of-the-art models in predicting diabetes, particularly when applied to structured EHR data with longer sequence lengths.
    \item We show that fine-tuning improves the performance of the clinical prediction model without adding additional tokens to the pre-trained tokenizer of the LLMs.
    \item We propose a method for post hoc explanation and disease risk assessment using LLMs combined with SHAP values to visualize textual laboratory test values.
    
\end{itemize}

The following sections of this paper are organized as follows. We summarize the limitations of machine learning (ML) techniques and briefly present the existing works applying LLMs in the healthcare domain in Section \ref{sec:related_work}. Our proposed approach is given in Section \ref{sec:method}. Section \ref{sec:data} provides an overview of data collection.  To demonstrate the effectiveness of our model, we conducted extensive experiments in Section \ref{sec:results}. Finally, we summarize the findings and implications of this work in Section \ref{sec:conclusion}.

\section{Related work}\label{sec:related_work}

\subsection{The Limitation of ML Methods}

This section explores the limitations of traditional ML methods, such as Logistic Regression and SVM, when applied to large-scale EHR data. These methods struggle with inherent challenges in EHR data, including missing values, imbalanced sample sizes, and the computational demands of processing large datasets. Although XGBoost's tree-based approach offers some mitigation for these issues, a major limitation of traditional ML methods remains: their inability to effectively model and predict diseases using diverse data types, such as text, images, and tabular data.

Predictive assessments in clinical settings are crucial for estimating a patient's risk of developing diseases, their potential response to treatment, and the likely course of their condition \cite{Laupacis}. Traditionally, models such as logistic regression \cite{Hosmer-2013} and random forest \cite{Breiman-2001} have been used for these disease prediction tasks. However, a key limitation of these approaches is their inability to effectively model the time-dependent nature of medical events, such as the order in which diagnoses, procedures, and medications occur.  Instead, they often simply focus on whether these events are present or absent as features, without considering their sequential importance.

\subsection{Large Language Models (LLMs)}
There are have been notable emphasis on LLMs within the AI domain. These models undergo extensive training using vast datasets and have shown impressive capabilities in various NLP applications, including language generation, machine translation, and question answering \cite{Han-2021}. Given the exponential expansion of medical literature and the growing accessibility of EHRs, LLMs are on the verge of transforming the field of medicine. LLMs have the potential to help healthcare professionals identify medical conditions \cite{Cascella-2023}. Examining patient information, including medical history and test results, these models can produce diagnoses and propose additional tests \cite{Chen-2023, Huang-2023, Kleesiek-2023}. This contributes to the reduction of diagnostic mistakes, streamlines the diagnostic procedure, and improves the overall standard of healthcare \cite{Chirino-2023}.

LLMs have the ability to revolutionize various aspects of medical practice, including improvements in diagnostic precision, forecasting disease progression, and aiding clinical decision making \cite{ChatCAD, MedBERT}. Through the analysis of extensive medical datasets, LLMs can quickly acquire specialized expertise in various medical fields, including radiology, pathology, and oncology \cite{RadBERT, Kather-2023}. These models can be refined using domain-specific medical literature, ensuring their currency and relevance. In addition, its adaptability to various languages and contexts promotes enhanced worldwide access to medical knowledge and expertise. 

\section{Method}\label{sec:method}

\subsection{Large Language Multimodal Models}
The majority of LLMs can be trained on large-scale text data before being applied as downstream models. However, most EHRs contain both numerical information (e.g., age, length of hospital stay, and laboratory test values) and categorical information. In our research, we explored two pre-training approaches: (1) we experimented with a multimodal approach that incorporates a text embedding encoder with multi-head attention learning on laboratory test values; (2) We converted laboratory test values of patients with chronic diseases into text information that can be used as input to LLMs. To facilitate effective learning of numerical blood features, we employed a simple DNN during the training process of the blood representation. We then concatenated the LLMs with the embedded blood feature. As inputs to our LLMMs, we selected representative laboratory test items for chronic diseases, as detailed in the appendix \ref{appendix:data}. This allowed us to achieve predictions that closely corresponded to clinical scenarios.

\subsection{Pre-training LLMs with Large Clinical Notes}
In our study, we initially developed a pre-training approach to extract text feature embedding from the EHR corpus. In Fig. \ref{fig:model}, we conducted pre-training of extensive EHRs using LLMs to acquire text representations. Our objective is to accomplish fusion learning by integrating the acquired text embeddings with patients who have concurrent laboratory test values. We performed various tokenization and pre-training techniques on our FEMH corpus, allowing the model to comprehend a significant amount of domain-specific clinical knowledge and contextual semantics. Subsequently, we engage in discussions with expert doctors to fine-tune the model. We evaluated the model in a large clinical corpus using LLMM training, which used basic BERT \cite{bert}, RoBERTa \cite{roberta}, ClinicalBERT \cite{alsentzer2019publicly}, BiomedBERT \cite{pubmedbert}, Flan-T5 \cite{flan-t5} and SciFive\cite{phan2021scifive} as extractors in order to reduce the number of model parameters and complexity during training. Furthermore, we used larger language models such as GPT-2 \cite{gpt2} for pre-training.

\subsection{Large laboratory Test Values Encoding}

To assess the risk of chronic diseases and improve our LLMM, we selected training data that included concurrent clinical notes and laboratory test samples from patients. Focusing on representative laboratory test items associated with diabetes and other chronic conditions, we establish the following definition of diabetes.

\begin{itemize}
    \item Fasting Plasma Glucose (FPG)  $\geqslant$ 126 mg/dL
    \item Glycated Hemoglobin (HbA1c) $\geqslant$ 6.5\%
\end{itemize}


We then designed a DNN sub-model to extract features from the laboratory blood samples. This sub-model was integrated with the LLMMs, which fused the extracted features with the semantic information from the chronic disease corpus within a latent space.

\subsection{Multi-head Attention Fusion}

In our overall framework, we concatenated two embeddings, which are text representation from LLM encoders and blood representation from the DNN's outputs. Thus, we have designed a multi-head attention module to facilitate a better fusion of features from the two domains in the latent space. This allows embedding vectors of text and blood features to perform a dot-product operation through the attention mechanism. We concatenated embedding vectors as query, key, and value for the attention module to generate attention-weighted matrices. By comparing the relevance of a query and key, attention weights determine the importance of each value in answering the current query, where a higher attention weight indicates a greater significance of each value for the query's resolution. After that, to enhance latent feature fusion, we utilized the final concatenated encoded features from multi-head attention embedding with LLMs and DNN vectors for final Multilayer Perceptron (MLP) layers. In Section \ref{sec:shap}, we performed an interpretable method to explain and highlight the risk text in the clinical area.

\subsection{Conversion of Laboratory Test Values to Text}

\begin{figure}
\centering
\includegraphics[width=1\textwidth]{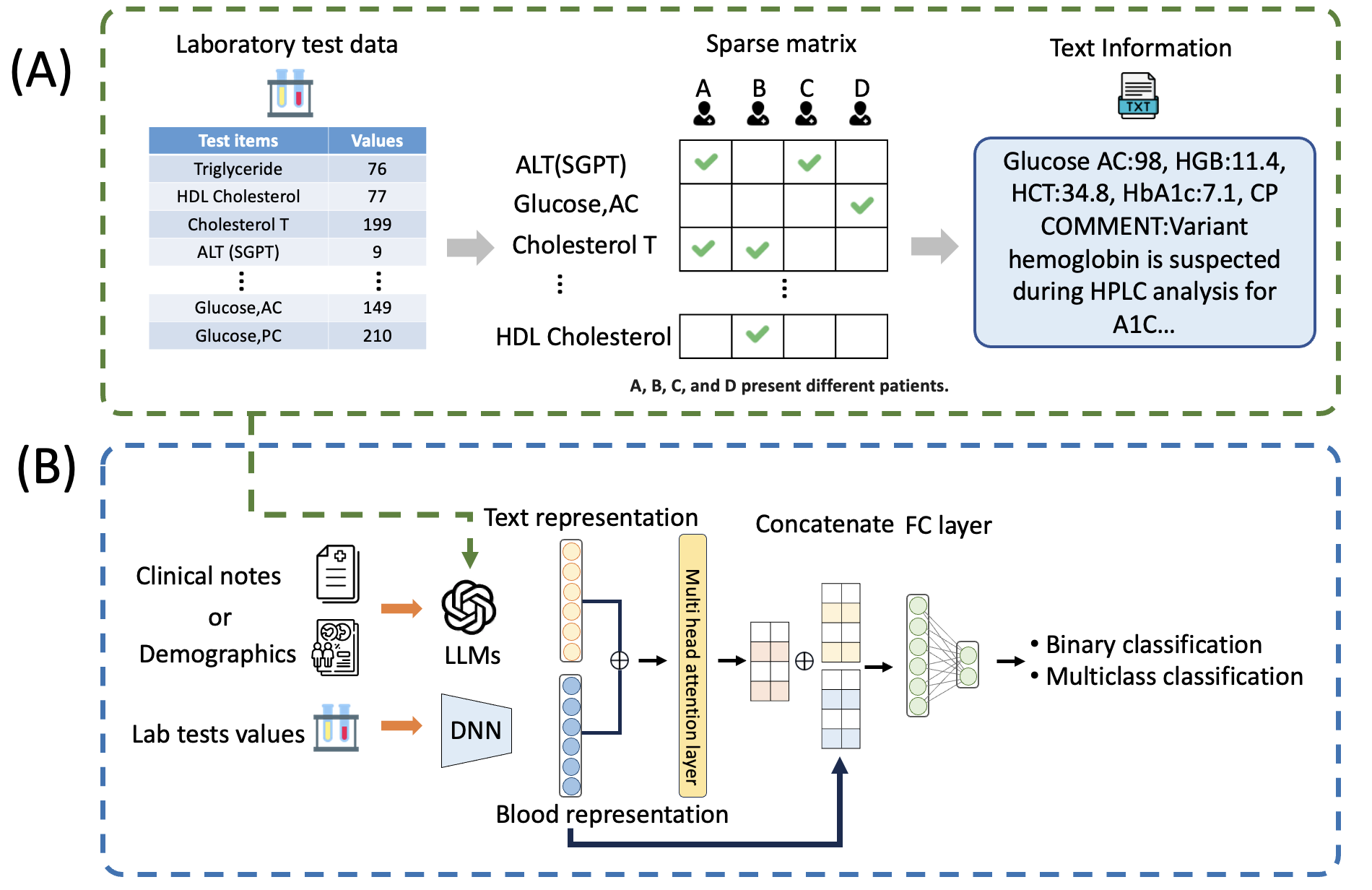}
\caption{The overall of our proposed LLMMs model framework. LLMMs will fuse the embedding and attention module of the language model with the 5-year laboratory test data. Figure (A) shows the language model training after the laboratory test values are textualized. Figure (B) shows the multimodal combination of numerical and laboratory test data.}
\label{fig:model}
\end{figure}

In most clinical settings, laboratory blood tests are routinely used to assess the risk of chronic disease. However, these tests may for the same patient may vary between measurements taken, leading to sparsity issue of laboratory values illustrated in Fig. \ref{fig:model}. To address this challenge, we converted laboratory test values into textual information. This approach mitigates the issue of sparse data in testing items and overcomes the limitation of LLMs in predicting textual outcomes from solely numerical features. In addition, it helps address the scenario in which patients might have missing data for most of their laboratory test items.

\section{Cohort Data Collection}\label{sec:data}

Different clinical notes involve making predictions or decisions based on various sources of clinical documentation, including doctor's notes, laboratory test results, and other medical records. Most of the NLP research in the field of EHR has utilized open datasets, such as the MIMIC series collected by the Medical Information Mart for Intensive Care (MGH). However, the majority of MIMIC data are limited by the small sample size and do not adequately represent the diversity needed for NLP clinics and training tasks. Therefore, our proposed approach used diverse real-world medical datasets from Far Eastern Memorial Hospital (FEMH) in Taiwan. Each dataset was divided into a 80\% training set and a 20\% testing set. Table \ref{table:medical-data} illustrates a brief overview of our input data format, detailing clinical notes, laboratory values, and textual laboratory values. The remaining group consists of numerical data that explicitly include items related to laboratory test results. We provide statistical samples of the dataset for the diabetes prediction task in Table \ref{tab:diabetes_data}.  

\begin{table}
\centering
\resizebox{\textwidth}{!}{
\begin{tabular}{|p{1.5cm}|p{3.5cm}|p{3.5cm}|p{7cm}|}
\hline
\textbf{Data} & \textbf{Data type} & \textbf{Description} & \textbf{Example} \\
\hline
{Corpus} & Clinical note & The outpatient physician records the patient'ss past records, including the patient's medical history, patient symptoms, and physician assessment. & This is a 56-year-old male patient with underlying hypertension and gout, with ophthalmic history of left eye retinal detachment status post pars plana vitrectomy with encircling buckling in 2016. This time, he complained of right eye with floaters since 2020.10, and he came to our and retinal breaks of right eye was noted. Therefore, a focal laser was applied to him. However, the patient complained of progressive visual field defect and blurred vision of right eye in recent days. Upon examination, decreased right eye vision (0.1) was found. Fundus examination showed retinal detachment from the upper parts with breaks was observed. Pars plana vitrectomy was suggested. He received surgery on 2020.12.24. After surgery, he was admitted for further treatment. \\
\cline{2-4}
 & Textual laboratory values & The laboratory examination section comprises the text description of laboratory items and the corresponding value. & Free T4:1.42, TSH:1.450, HDL Cholesterol:57, BUN:22, Cholesterol T:146, Estimated GFR(MDRD):60, Glucose AC:148, ALT (SGPT):29, Uric Acid:6.5, Creatinine:1.22, K:4.5, Triglyceride:66, LDL Cholesterol:88 \\
\hline
Numerical data & Blood test items & \multicolumn{2}{p{10cm}|}{Record the biochemical data and test indicators. }  \\
\hline
\end{tabular}}
\caption{A description of the LLMMs' training corpus and textual laboratory value.}
\label{table:medical-data}
\end{table}

\begin{table}
\centering
\resizebox{\textwidth}{!}{
\begin{tabular}{|l|l|l|c|c|c|c|}
\hline
\textbf{Disease} & \textbf{Task} & \textbf{Samples define}  & \textbf{Training size} & \textbf{\#Diabetes} & \textbf{\#Non-diabetes} \\ \hline
Diabetes & Clinical notes classification & Has a history of diabetes. & \textbf{212,936} & \textbf{52,458} & \textbf{160,478} \\ \cline{2-6}
 & Laboratory values to text classification& Initial onset of diabetes ( < 180 days) & \textbf{1,750,711} & \textbf{7,892} & \textbf{1,742,879} \\ \cline{2-6}
 & \begin{tabular}[c]{@{}l@{}}Text + Numerical features \\ prediction\end{tabular}& \{Diabetes disease $\cap$ laboratory test data\} & \textbf{21,683} & \textbf{6,844} & \textbf{14,839} \\ \hline
\end{tabular}}
\caption{Summary of Diabetes Prediction Data.}
\label{tab:diabetes_data}
\end{table}

\section{Experiments}\label{sec:results}

\subsection{Baselines}

We evaluated our model performance with ML-based models, such as XGboost \cite{chen2016xgboost} and DNN based on only numerical laboratory test values. Besides, LLMMs with multimodal fusion used different and classic language models as the backbone to learn the embedding of the corpus, and at the same time combined the attention module to fuse the laboratory test values. We selected classical LLMs as the backbone for our LLMMs, including BERT, Flan-T5-base-220M, Flan-T5-large-220M, GPT-2, SciFive, RoBERTa, BiomedBERT, and ClinicalBERT as our baseline models. Building upon this foundation, we then developed a multimodal system by incorporating an additional attention fusion module that integrates information from different modalities.

\begin{table}
\centering
\caption{Overall model comparison}
\resizebox{\textwidth}{!}{
\begin{tabular}{lcccccc}
\hline
\multicolumn{7}{c}{\textbf{Diabetes prediction}} \\
\hline
\textbf{Models} & \textbf{Text (Clinical notes)} & \textbf{Lab test values} & \textbf{Accuracy} & \textbf{Recall} & \textbf{Precision} & \textbf{F1-score} \\
\hline
\hline
Flan-T5-base-220M & \textbf{$\checkmark$} & - & 0.95 & 0.95 & 0.95 & \textbf{0.95} \\
SciFive & $\checkmark$ & - & 0.93 & 0.91 & 0.94 & 0.93 \\
RoBERTa & $\checkmark$ & - & 0.87 & 0.87 & 0.87 & 0.87 \\
BiomedBERT + attention & $\checkmark$ & $\checkmark$ & 0.92 & 0.92 & 0.91 & 0.91 \\
RoBERTa + attention & $\checkmark$ & $\checkmark$ & 0.91 & 0.91 & 0.91 & 0.91 \\
SciFive + attention & $\checkmark$ & $\checkmark$ & 0.90 & 0.90 & 0.90 & 0.90 \\
ClinicalBERT + attention & $\checkmark$ & $\checkmark$ & 0.90 & 0.90 & 0.90 & 0.90 \\
GPT-2 + attention & $\checkmark$ & $\checkmark$ &  0.88 & 0.88 & 0.89 & 0.88 \\
BERT + attention & $\checkmark$ & $\checkmark$ & 0.84 & 0.85 & 0.84 & 0.84 \\
XGBoost & - & $\checkmark$ & 0.84 & 0.82 & 0.82 & 0.81 \\
DNN & - & $\checkmark$ & 0.78 & 0.78 & 0.77 & 0.77 \\
\hline
\multicolumn{7}{c}{\textbf{Multiclass prediction}} \\
\hline
\hline
SciFive & $\checkmark$ & - & 0.72 & 0.72 & 0.71 & 0.71 \\
RoBERTa & $\checkmark$ & - & 0.68 & 0.68 & 0.66 & 0.66 \\
Flan-T5-large-770M & $\checkmark$ & - & 0.73 & 0.73 & 0.72 & 0.72 \\
ClinicalBERT + attention & \textbf{$\checkmark$} & \textbf{$\checkmark$} & 0.73 & 0.73 & 0.73 & 0.73 \\
BiomedBERT + attention & $\checkmark$ & $\checkmark$ & 0.73 & 0.73 & 0.72 & 0.72 \\
RoBERTa + attention & $\checkmark$ & $\checkmark$ & 0.72 & 0.72 & 0.71 & 0.71 \\
SciFive + attention & $\checkmark$ & $\checkmark$ & 0.71 & 0.73 & 0.70 & 0.70 \\
Flan-T5-base-220M + attention & $\checkmark$ & $\checkmark$ & 0.71 & 0.71 & 0.70 & 0.70 \\
GPT-2 + attention & $\checkmark$ & $\checkmark$ &  0.71 & 0.71 & 0.69 & 0.70 \\
Flan-T5-large-770M + attention & $\checkmark$ & $\checkmark$ & 0.70 & 0.70 & 0.70 & 0.7 \\
BERT + attention & $\checkmark$ & $\checkmark$ & 0.59 & 0.58 & 0.57 & 0.58 \\
XGBoost & - & $\checkmark$ & 0.54 & 0.37 & 0.37 & 0.37 \\
DNN & - & $\checkmark$ & 0.50 & 0.50 & 0.46 & 0.44 \\
\hline
\end{tabular}}
\label{tab:model_comparison}
\end{table}

\subsection{Early Diabetes and Multiclass Chronic Disease Prediction}

In this section, we explore two vital prediction tasks in healthcare and medical research through our LLMMs model, focusing on multiclass chronic disease prediction and early diabetes prediction as primary multimodal applications in the clinical domain. First, we introduced early prediction of diabetes to identify people at risk of developing diabetes before the appearance of clinical symptoms. This task aimed to assess the likelihood of future diabetes onset using predictive models based on relevant clinical notes or laboratory test values, formulated as a binary classification problem. The second objective was the prediction of multiclass chronic diseases, which assesses the probability that an individual develops one of several chronic diseases or conditions. In the multiclass prediction, we classified into five categories, including diabetes with hypertension, diabetes with hyperlipidemia, hypertension alone, and other types. 

We evaluate the performance of our LLMMs on modality data, as shown in Table \ref{tab:model_comparison}. In our multimodal attention fusion experiment, BiomedBERT achieved the highest performance, with an accuracy of 0.92 and an F1 score of 0.91 for diabetes prediction. Our experimental results demonstrate that incorporating an attention model into the model using both clinical notes and laboratory values leads to more accurate prediction of patients with early diabetes compared to use laboratory values only. In the classification of multiclass chronic diseases, ClinicalBERT achieved an accuracy and F1 score of 0.73. When comparing different models for chronic diseases classification, we found that models relying solely on clinical notes struggle with model explainability for early disease prediction. This contrasts with traditional methods like XGboost and DNN that use laboratory values, as well as newer methods like SciFive and RoBERTa that only use clinical notes. By validating the numerical report, we can enhance the overall model in the latent space and improve its fusion with textual text embedding. However, in our observations, we noticed that larger language models like T5-large, and GPT-2 didn’t yield superior classification results. This suggests that language models may perform better when tailored specifically for medical domain tasks.

\subsection{Early Diabetes Prediction by Textual Laboratory Test Values}

\begin{figure}
\centering
\includegraphics[width=1\textwidth]{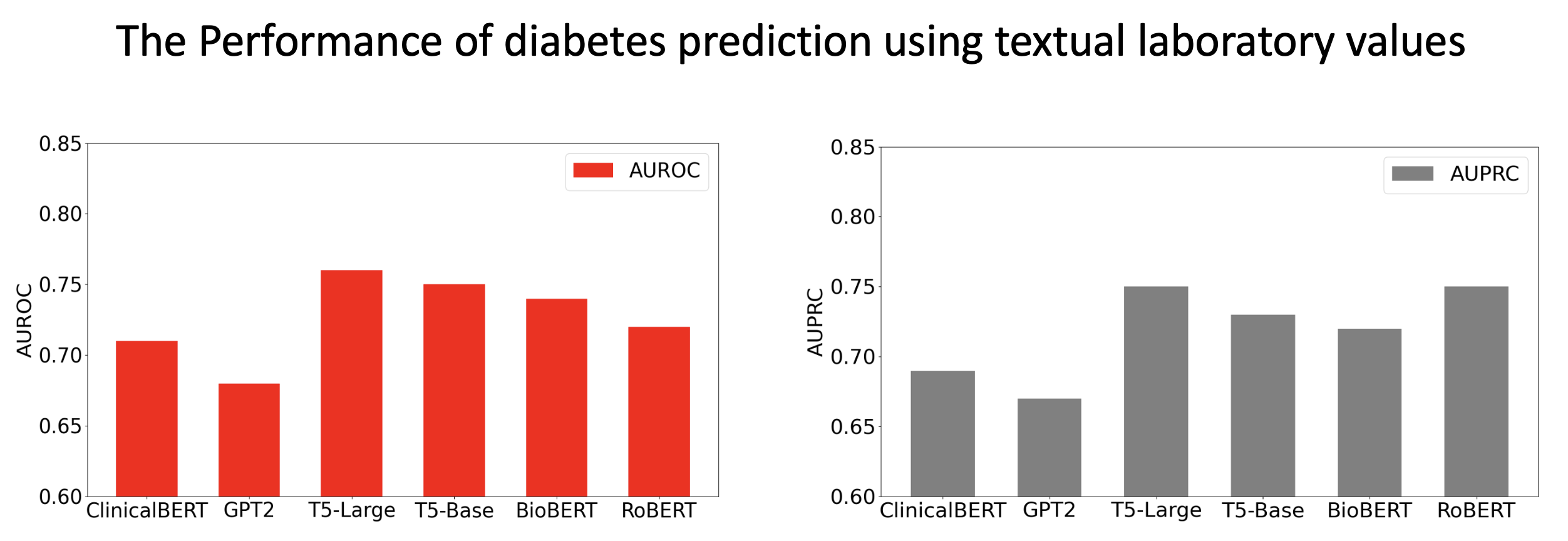}
\caption{AUC vs. AUPRC performance of LLMMs using different pre-train models at 5 years of textual laboratory blood value corpus.}
\label{fig:textual_lab_values}
\end{figure}

In the traditional perspective, disease modeling primarily relies on numerical laboratory values data. We strictly follow clinical physician guidance to filter and select patients based on laboratory blood values, identifying those who developed diabetes prior to their initial diagnosis. Subsequently, we extract laboratory blood values from this group of patients for textual encoding, facilitating more straightforward corpus encoding for our LLMMs. In Fig. \ref{fig:textual_lab_values}, it is evident that through the training of T5-large on textual data, the model achieved an AUC of 0.76 and an AUPRC of 0.75. Our experimental results indicate that solely using laboratory blood values leads to two issues: lower positive rates and the missing tests for most patients. Through textualized laboratory values, LLMs enable us to more accurately predict important interpretive laboratory test items related to newly diagnosed diabetes patients.

\subsection{Interpretable Attention in Contextual Laboratory Test Values}\label{sec:shap}

\begin{figure}
\centering
\includegraphics[width=1\textwidth,height=1\textheight,keepaspectratio]{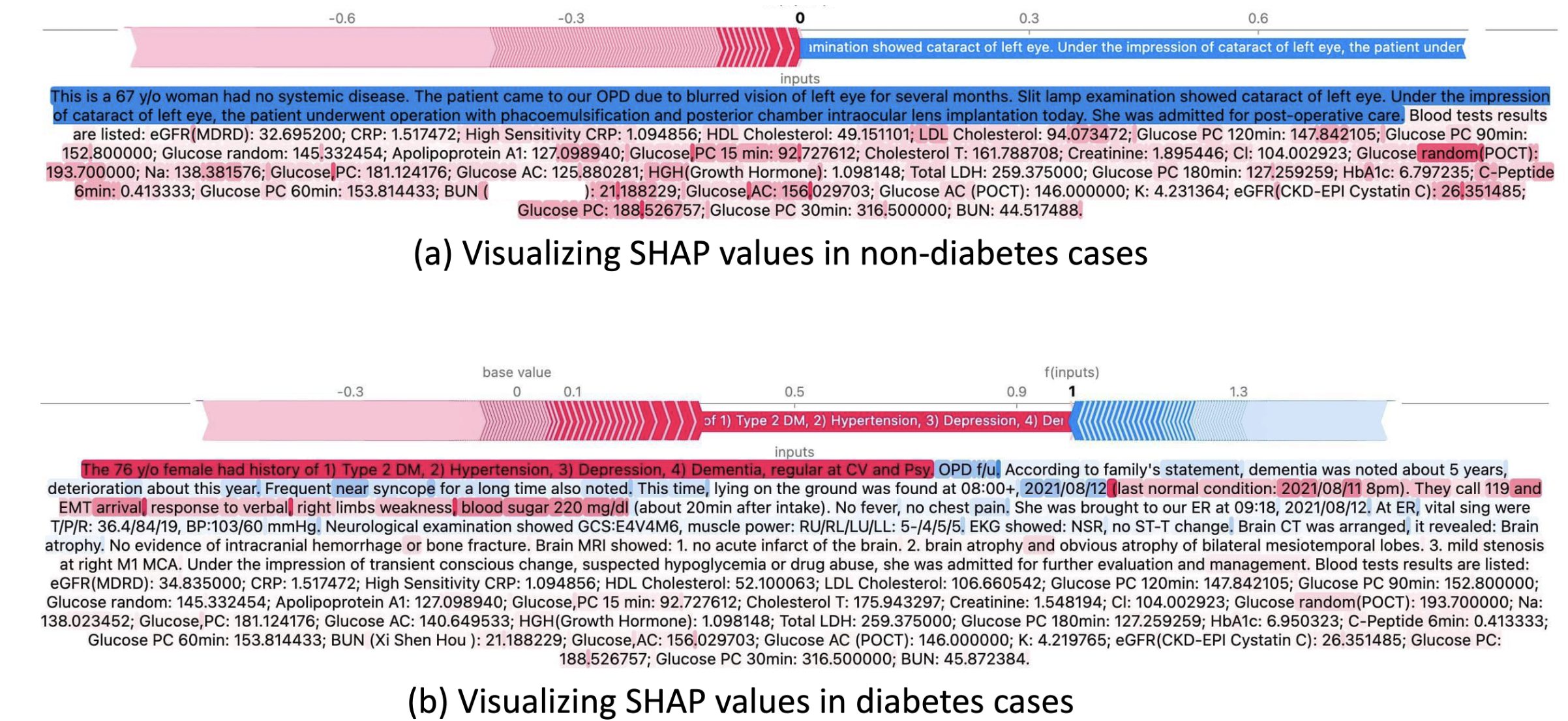}
\caption{Comparing the interpretation sample of both cases diabetics and non-diabetic with Shapley Value}
\label{fig:shap}
\end{figure}

In this section, we applied the Shapley values by calculating the attention scores LLMs and visualizing the highlighted high-risk text (laboratory test results and clinical notes) based on the attention scores. The visual representation provides a comparative analysis of two distinct sample sets used in diabetes prediction. In Fig. \ref{fig:shap}, the top figure illustrates a sample from non-diabetic individuals, analyzed using Shapley values. These values serve as a powerful tool for understanding the individual contributions of each input variable to the predictive model's output. Specifically, they reveal the complex interplay between various clinical indicators and predicted outcomes, including important markers like glucose and Glucose AC levels, which play a crucial role in diagnosing and managing diabetes. In contrast, the bottom figure employs a novel approach to model interpretation for a diabetic sample. Here, the focus shifts towards leveraging the power of NLP to extract key insights from textual clinical notes. By identifying and analyzing critical keywords within the narrative, the model sheds light on the intricate relationships between textual data and diabetic outcomes, offering a comprehensive understanding of the predictive process.

\section{Conclusion}\label{sec:conclusion}

In this research, we validate the feasibility of LLMs on large-scale clinical notes and introduce an attention mechanism to better fuse clinical notes with laboratory test values. Specifically, through different pre-trained language models, we observe that clinicalBERT and BiomedBERT with attention fusion can achieve better prediction results for multiclass chronic diseases and single diabetes. At the same time, LLMs with attention modules can also provide interpretable clinical notes using Shapley values for textual laboratory test values. In the future, our proposed model can provide a real-time and effective risk warning system for clinicians and patients.

\appendix
\section{Appendix}
\subsection{Details of cohort}\label{appendix:data}

\begin{itemize}
\item \textbf{Blood test items}: eGFR (MDRD), CRP, High Sensitivity CRP, HDL Cholesterol, LDL Cholesterol, Glucose PC 120min, Glucose PC 90min, Glucose random, Apolipoprotein A1, Glucose PC 15 min, Cholesterol T, Creatinine, Glucose random (POCT), Na, Glucose PC, Glucose AC, HGH (Growth Hormone), Total LDH, Glucose PC 180min, HbA1c, C-Peptide 6min, Glucose PC 60min, BUN, Glucose AC (POCT), K, eGFR (CKD-EPI Cystatin C), Glucose PC 30min, Creatinine (POCT), ALT (SGPT), AST (SGOT), Triglyceride
\end{itemize}

\bibliographystyle{unsrt}  
\bibliography{references}

\end{document}